\documentclass[10pt,twocolumn,letterpaper]{article}
\usepackage{preprint}
\usepackage[T1]{fontenc}
\usepackage{DejaVuSans}
\usepackage{colortbl}
\definecolor{flinchtint}{HTML}{FFF0E2}
\newcommand{\figlabelsize}{8}
\newcommand{\figtableleading}{10}

\definecolor{linkblue}{rgb}{0.21,0.49,0.74}
\usepackage[breaklinks,colorlinks,allcolors=linkblue]{hyperref}
\usepackage[capitalize]{cleveref}
\crefname{section}{Sec.}{Secs.}
\Crefname{section}{Section}{Sections}
\crefname{appendix}{Sec.}{Secs.}
\Crefname{appendix}{Section}{Sections}
\crefname{table}{Tab.}{Tabs.}
\Crefname{table}{Table}{Tables}
\newcommand{\name}{FLINCH\xspace}
\newcolumntype{Y}{>{\raggedright\arraybackslash}X}
\newcolumntype{L}[1]{>{\raggedright\arraybackslash}p{#1}}
\hypersetup{
  pdftitle={FLINCH: Preserving Subtle to Large Facial Expressions in 3D Face Models},
  pdfauthor={Swethasree Sribabu, Vishnu Raj, Gouthaman KV, Manivannan Muniyandi},
  pdfcreator={LaTeX}
}

\title{FLINCH: Preserving Subtle to Large Facial Expressions in 3D Face Models}
\author{
  Swethasree Sribabu$^{1}$\quad
  Vishnu Raj$^{2}$\quad
  Gouthaman KV$^{2}$\quad
  Manivannan Muniyandi$^{1}$\\[0.4em]
  $^{1}$Indian Institute of Technology Madras\qquad
  $^{2}$Dolby Laboratories\\[0.2em]
  \small am24s015@smail.iitm.ac.in,\quad
  \{vishnu.raj, gouthaman.kv\}@dolby.com,\quad
  mani@iitm.ac.in
}
\date{}
\newcommand{\projectpageline}{}

\newcommand{\preprintteaser}{
  \begin{center}
    \includegraphics[width=\textwidth]{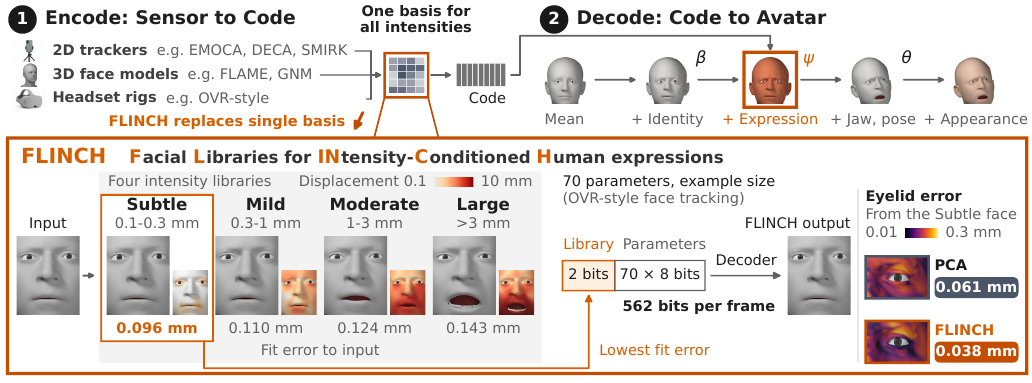}
    \captionof{figure}{\protect\textbf{FLINCH (Facial Libraries for INtensity-Conditioned Human expressions) replaces one expression basis with four libraries, one per intensity band.} Intensity bands are set by RMS displacement from the neutral face, with each band's limits in mm shown above it. Top: a typical avatar pipeline codes every expression with one basis dominated by large expressions, losing low-intensity ones. Bottom: FLINCH codes one input frame. All faces are real, registered FaMoS~\cite{famos2023} frames of one person whose data were not used to build the libraries. Each library shows an example face from its range and that face's displacement map. Every library reconstructs the input, and the one with the lowest error is chosen. This input lies in the Subtle band (0.26 mm), and the Subtle library fits it best (0.096 mm). Its 2-bit index and 70 parameters (close to an OVR-style code size) take 562 bits, only 2 bits more than one basis. Right: the Subtle example face's eye, coded by PCA and FLINCH at the same size. FLINCH's RMS error is 37\% lower. Icons are generated. $\beta$, $\psi$ and $\theta$ are FLAME's shape, expression and pose parameters.
}
    \label{fig:schematic}\label{fig:teaser}
  \end{center}
}

\begin{document}
\maketitle
\thispagestyle{plain}
\begin{abstract}
Low-intensity facial expressions move the face by less than a millimetre on average. A slight tightening of the eyelids is one example. Such expressions make avatars look alive in video calls, games, film and augmented and virtual reality. Avatar pipelines store each frame as a short expression code, and face-tracking headsets send every expression through it. This code is fitted to average squared error, so large expressions dominate it and low-intensity ones are lost. We present FLINCH (Facial Libraries for INtensity-Conditioned Human expressions), an intensity-conditioned expression representation. It has one shape library per intensity band of root-mean-square (RMS) displacement from the neutral face: Subtle (0.1 to 0.3\,mm), Mild (0.3 to 1\,mm), Moderate (1 to 3\,mm) and Large (above 3\,mm). Each frame is stored as the parameters of whichever library reconstructs it best, plus a 2-bit index. Decoding stays linear, and the code is only 2 bits longer than a standard code with the same number of parameters. Evaluating each band separately reveals the loss that a single average error hides. On the FaMoS dataset at 70 parameters, FLINCH has 53\% lower squared error than PCA trained on the same data on Subtle expressions and 26\% lower on Mild ones, with no measurable loss on Large ones. PCA needs 117 parameters to match FLINCH on Subtle expressions. FLINCH preserves Subtle expressions better at every code size tested, from 8 to 383 parameters. It also has lower error than the expression spaces of FLAME and GNM, a VR headset rig and mesh autoencoders. It keeps 99\% of a Subtle eyelid movement, against 86\% for FLAME 2023. Refitted on the Ava-256 dataset, FLINCH again has lower error than PCA in every band. It also lowers error on real spontaneous micro-expressions captured in 4D (4DME). We also build FLINCH-4D, a benchmark of over a million Ava-256 frames labelled by expression intensity, for release.
\end{abstract}

\section{Introduction}
\label{sec:intro}
People are remarkably sensitive to faces. A slight tightening of the eyelids or a small change at the mouth corner can alter how an expression is perceived~\cite{emoca2022}. These details are what make an avatar in virtual reality, a video call, a game or a film feel alive. Face tracking is becoming common in augmented and virtual reality headsets~\cite{bai2024}, so avatars will increasingly carry people into telepresence meetings, remote therapy, medical training, education and live entertainment, where a hesitant smile can carry meaning. Before an avatar can show an expression, the expression must be compressed into a compact code. Headset trackers, video-call models and film rigs all reduce each frame to a short vector of expression parameters~\cite{lewis2014blendshapes,bai2024}. Anything this code cannot represent is lost, and no tracker or renderer downstream can faithfully restore it (\cref{fig:grid}).

Expression capture has advanced quickly, from images~\cite{emoca2022,smirk2024} to headset cameras~\cite{bai2024}. Every captured expression, however, still passes through such a code, for example the expression basis of FLAME, and this code loses the most on low-intensity expressions. We divide expressions into four intensity bands by how far they move the face on average, Subtle (0.1 to 0.3\,mm), Mild (0.3 to 1\,mm), Moderate (1 to 3\,mm) and Large (above 3\,mm), and name each expression by its band. Low-intensity expressions are the Subtle and Mild ones. Such expressions are far from rare. In spontaneous headset recordings of people suppressing their expressions~\cite{blendme2026}, 69\% of the frames that show an expression show a low-intensity one (95\% interval 59 to 79\%, measured through a headset blendshape rig as described in the supplement). Even when 36 people pose expressions on cue in a virtual reality game~\cite{ortmann2024emojiherovr}, 22\% of these frames still show a low-intensity expression (18 to 26\%).

Expression codes lose low-intensity expressions because of how they are learned. PCA models and the learned expression spaces of morphable models such as FLAME~\cite{flame2017} are fitted to minimize the average reconstruction error over all training expressions. We treat an expression as the non-rigid change of the face from its neutral shape after head pose is removed, and its intensity as how far the face moves, measured as the root-mean-square (RMS) displacement in millimetres. Squared error grows quickly with size: an error of 3\,mm weighs a hundred times more than an error of 0.3\,mm. Fitting a code to minimize this average therefore focuses it on large expressions. A single basis, the fixed set of face-shape directions that the code combines, uses most of its directions to describe them, as the energy-maximising property of PCA implies~\cite{fan1949theorem}. The resulting model achieves a low average error while discarding much of the detail in low-intensity expressions, a loss that the single average score conceals. Our key observation is that the same number of parameters can serve every intensity if it is divided among intensities rather than shared by one basis.

We therefore introduce \name (Facial Libraries for INtensity-Conditioned Human expressions), an intensity-conditioned expression representation for 3D face models that gives every intensity its own share of the parameters. Instead of one expression basis, \name learns four shape libraries (\cref{fig:schematic}), one for each intensity band: Subtle, Mild, Moderate and Large. Each library is a small set of face-shape directions learned from expressions in its band. For each frame, \name chooses whichever library reconstructs it best and sends its parameters with a 2-bit index. This adds almost no cost: the code is only 2 bits longer than a standard code of the same size, and decoding remains one matrix multiplication, as fast as with a standard basis. Any pipeline that carries blendshape-style parameters can therefore carry \name. \name also works at every code size we tested, from 8 to 383 parameters, and we report 70, close to the size of an OVR-style face-tracking code. It selects the library from the 3D displacement, so it applies directly wherever a mesh is available, such as in capture and animation pipelines. From images, a tracker trained on rendered faces drives \name with an error within 0.3\% of PCA's (\cref{sec:real}). Mixtures of subspaces~\cite{tipping1999mixtures} are the closest prior idea. At 70 parameters, \name has 8 to 39\% lower squared error on Subtle expressions than standard mixtures, and a mixture comes within 3\% only when it is given intensity labels (\cref{sec:mixture}). On FaMoS 4D scans~\cite{famos2023} at 70 parameters, \name lowers squared error by 53, 26 and 10\% on Subtle, Mild and Moderate expressions against PCA trained on the same data, with no measurable loss on Large ones (\cref{sec:main}). It preserves Subtle expressions better at all nine tested sizes. A single average reconstruction error would hide these differences, so we report error separately for each intensity band and introduce FLINCH-4D, a benchmark that labels every frame by expression intensity.

Our contributions are:
\begin{itemize}
\item \textbf{Intensity-conditioned expression libraries (\name).} We introduce an intensity-conditioned expression representation that replaces the expression basis of a 3D face model. Its decoding stays linear, it adds only 2 bits per frame, and on FaMoS at 70 parameters it preserves Subtle expressions better than every alternative we test, including PCA trained on the same data, FLAME, a headset rig, mesh autoencoders, subspace mixtures and GNM~\cite{gnm2026}, which uses more than five times as many parameters.
\item \textbf{Evaluation of expression models by intensity bands.} We show that a single average error hides the loss of low-intensity expressions. Per-band evaluation reveals losses in facial-action peaks, expression classification and mixed expressions, and shows that a basis fitted to average error suffices only for Large expressions.
\item \textbf{FLINCH-4D benchmark.} We build a 4D benchmark that labels every frame by expression intensity. It contains over a million registered Ava-256 frames from 213 people, with whole-face and regional intensity labels, the start, peak and end frames of each expression, and training and test sets drawn from different people. It will be released.
\end{itemize}

Together, \name and FLINCH-4D make intensity a dimension that expression models are measured on and designed for, so that the low-intensity expressions that make an avatar feel alive survive compression.

\section{Related Work}
\label{sec:related}
\paragraph{Expression representations.}
Among morphable face models~\cite{egger2020survey}, LSFM~\cite{lsfm2016} and UHM~\cite{uhm2020} model identity and head shape, and FaceWarehouse~\cite{cao2014facewarehouse}, FLAME~\cite{flame2017} and GNM~\cite{gnm2026} add expression spaces. Region PCA~\cite{tena2011} and SPLOCS~\cite{splocs2013} model local deformation, and mesh autoencoders learn nonlinear codes~\cite{coma2018,gong2019spiralnet}. Neural parametric heads~\cite{nphm2023,imfacepp2025,serep2025} and local avatar models~\cite{scaffoldavatar2025,expportrait2026} add finer expression detail. Blendshape rigs~\cite{lewis2014blendshapes,arkitblendshapes} support editing, compressed skinning~\cite{kavan2024}, Gaussian blendshapes~\cite{ma2024gaussianblendshapes}, GRMM~\cite{grmm2025} and GaussianAvatars~\cite{gaussianavatars2024} improve the storage, appearance or detail of rigged heads, and expression cloning~\cite{noh2001expression} and deformation transfer~\cite{sumner2004transfer} retarget expressions between faces. Local PCA~\cite{kambhatla1997dimension}, mixtures of probabilistic PCA~\cite{tipping1999mixtures} and subspace clustering~\cite{vidal2011subspace} use several subspaces, and parameterized PCA lets subspaces depend on a known context variable~\cite{gupta2018parameterized}. These models fit one representation to all intensities or split it without regard to intensity. In contrast, \name assigns one library to each intensity band and chooses whichever library reconstructs a frame best (\cref{sec:main}).

\paragraph{Expression capture from images and headsets.}
DECA~\cite{deca2021}, EMOCA~\cite{emoca2022}, SMIRK~\cite{smirk2024}, Pixel3DMM~\cite{pixel3dmm2025} and FlowFace~\cite{taubner2024flowface} recover 3D expressions from images and video, building on dense face alignment~\cite{guler2017densereg,jeni2017dense}, and Sun et al.~\cite{micro2026} reconstruct micro-expressions from global dynamics and local optical-flow, landmark and mesh cues. Motion magnification~\cite{wu2012eulerian} and MagPlus~\cite{jammal2026magplus} amplify subtle expressions to make them visible, whereas \name preserves them in the code. FaceVR~\cite{facevr2018} and Song et al.~\cite{song2018} track faces under headset occlusion, image translation~\cite{wei2019vr} and Codec Avatar encoders~\cite{bai2024} map headset views to avatars, OFERA~\cite{ofera2026} drives Gaussian avatars from occluded headset blendshapes, and synthetic headset views support avatar training~\cite{genhmc2025} and expression recognition~\cite{synhmc2026}. Synthetic data also supports face analysis~\cite{wood2021}, expression augmentation~\cite{smirk2024}, micro-expression synthesis~\cite{miex2022} and emotional speech animation~\cite{emote2023}. BlendME~\cite{blendme2026} records headset blendshape micro-expressions with retrospective self-annotation. These methods recover expressions ever more accurately, but each outputs them as a fixed expression code, often FLAME parameters. \name changes that code rather than the method that predicts it, so it complements them. For a fair comparison of the codes themselves, our experiments give every expression space the same 3D input.

\paragraph{Intensity and 4D expression data.}
CASME II~\cite{casme2}, SMIC~\cite{smic2013} and SAMM~\cite{samm2018} provide spontaneous micro-expression video, CAS(ME)$^3$~\cite{casme3} adds depth, 4DFAB~\cite{cheng2018fourdfab} and 4DME~\cite{fourDME2023} add 4D capture, and MEGC2025~\cite{megc2025} evaluates spotting, recognition and visual question answering. FACS~\cite{facs2002} defines action-unit coding, Mavadati et al.'s DISFA~\cite{disfa2013} labels action-unit intensity, and BP4D~\cite{bp4d2014} supplies spontaneous 3D sequences with FACS annotation. MEAD~\cite{mead2020} labels performed emotional intensity at three levels besides neutral, and Montanha et al.~\cite{montanha2025micromacro} study how viewers perceive micro- and macro-expressions on driven virtual humans. Intensity has also been used to control animation~\cite{medtalk2025} and to split evaluation~\cite{fu2026expressiondiverse}. Ava-256~\cite{ava2562024}, FaMoS~\cite{famos2023}, CoMA~\cite{coma2018} and D3DFACS~\cite{cosker2011d3dfacs} support geometric expression analysis, building on registered performance capture~\cite{klaudiny2012high}. These resources label emotions, action units or performed intensity rather than measured displacement. FLINCH-4D adds displacement-based intensity labels to over a million registered frames, and \name uses intensity to divide its parameters among the intensity bands.

\section{Intensity-conditioned libraries}
\label{sec:method}
\label{sec:overview}
For each frame, \name sends the $K$ parameters of the library that reconstructs it best, plus a 2-bit library index (\cref{fig:schematic}). A frame is the displacement of every registered mesh vertex from the neutral face (\cref{sec:intro}). Each library is learned from the expressions of one intensity band, so low-intensity expressions are coded with directions learned from expressions of their own intensity.

\subsection{Intensity and libraries}
\label{sec:code}
A frame holds the 3D displacements, in mm, of the $n=1{,}787$ vertices in the FLAME face mask~\cite{flame2017}, which leaves out the neck, scalp and ears. They form an $n\times3$ array, which we flatten into a vector $x\in\mathbb{R}^{3n}$. Its intensity $a(x)=\|x\|/\sqrt{n}$ is the RMS displacement per vertex. The four bands are Subtle $[0.1,0.3)$, Mild $[0.3,1)$, Moderate $[1,3)$ and Large, above $3$\,mm. Each band spans about three times the displacement of the one below. These band limits are rounded, and replacing them with exactly log-spaced limits changes the errors very little (supplement). Frames below $0.1$\,mm count as neutral and are not fitted. A run of consecutive non-neutral frames is an expression event. A strong but local expression, such as a blink while the rest of the face stays still, can have low whole-face intensity, so FLINCH-4D also labels regional intensity (\cref{sec:data}).

The jaw is coded the same way in every library. Three fixed orthonormal jaw directions, the columns of $J$, split a frame into its jaw part $JJ^\top x$ and the remaining expression $z=(I-JJ^\top)x$, where $I$ is the identity. The other $r=K-3$ parameters code $z$. For each band $b$, we fit PCA to the band's training expressions $z$ without subtracting their mean. The resulting basis $U_b$ is the band's library. Its $r$ columns are the directions that capture these expressions best in least squares. Without a mean, all-zero parameters give the neutral face, as all-zero weights do in a blendshape rig.

\subsection{Encoding a frame}
\label{sec:encoding}
For each library $b$, the encoder reconstructs the frame as
\begin{equation}
\hat{x}_b=JJ^\top x+U_bU_b^\top z,
\end{equation}
the jaw part plus the projection of the remaining expression onto library $b$. It keeps the library with the smallest whole-face reconstruction error $\|x-\hat{x}_b\|$ and sends the jaw parameters $J^\top x$, the expression parameters $U_b^\top z$ and the index of $b$. The choice of library depends on fit, not on the measured band, so a Mild frame may use the Subtle library. The decoder reads the index and forms $\hat{x}_b$. Selecting the best-fitting subspace and including its index in the code follows mixtures of probabilistic PCA~\cite{tipping1999mixtures}. Unlike a mixture, which learns its subspaces without labels, \name learns each library from the frames of one intensity band.
At $K=70$, the 67 expression and three jaw parameters take eight bits each, 560 bits in total, the same as a single basis with 70 parameters. The 2-bit index brings this to 562 bits per frame. 

\subsection{When is one basis enough?}
\label{sec:one-basis}
Separate libraries help only if intensity bands differ in their best directions. Let $S_b$ be the average of $zz^\top$ over band $b$'s training frames. Its eigenvectors are the directions that the band's expressions occupy, and its eigenvalues measure how strongly. Let $\pi_b>0$ be the band's share of training frames. Then $S=\sum_b\pi_bS_b$ is the average over all training frames, which a single PCA basis fits. Let $\phi_r(\cdot)$ be the sum of a matrix's $r$ largest eigenvalues, the energy that its best $r$-dimensional subspace captures. With the jaw directions fixed, separate band libraries lower the per-vertex training error of one shared basis by
\begin{equation}
G_J=\frac{\sum_b\pi_b\phi_r(S_b)-\phi_r(S)}{n}\geq 0.
\label{eq:kyfan}
\end{equation}
This is the energy captured by each band's best subspace minus that captured by the best shared one.

\textbf{Proposition.} $G_J=0$ if and only if the bands share an optimal $r$-dimensional subspace. Otherwise, band libraries strictly reduce training error. This follows from Ky Fan's maximum principle~\cite{fan1949theorem}. Choosing the best-fitting library for each frame can lower the error further and never raises it.

For instance, suppose every band's $S_b$ were one common matrix $S_0$ times a band factor $c_b$, plus noise of equal strength $\sigma_b^2$ in every jaw-free direction, $S_b=c_bS_0+\sigma_b^2(I-JJ^\top)$. The top directions would then coincide and $G_J=0$. If small expressions were scaled-down large ones, one basis would therefore be enough. The supplement tests this on Ava-256 expression events.

\paragraph{Why one basis loses Subtle expressions.}
Subtle expressions are small, so they add little to $S$, and a basis fitted to all frames stays close to the directions of larger expressions. Directions used mainly by Subtle expressions are therefore poorly represented. The supplement makes this precise with a perturbation bound, which at $K=70$ accounts for 81 to 83\% of PCA's excess error on Subtle expressions in the training folds. Eq.~\eqref{eq:kyfan} holds for any split of the frames, and \name fixes its split by intensity before training. At 70 parameters, it gives lower error than mixture models, which learn their own splits from the data (\cref{sec:main}).

\section{Experiments}
\label{sec:experiments}
\subsection{Data and protocol}
\label{sec:setup}
The FaMoS dataset~\cite{famos2023} provides registered 4D scans. Five folds over 75 people use 60 for fitting and 15 for validation. A further 20 people form the test set and are never used for fitting. Training keeps every Subtle frame and at most 128 frames from the other bands per person, with each sequence's first frame as the neutral face.
We measure error as the squared distance between the reconstructed and true displacement, averaged over the 1,787 vertices. We then average over the frames of each band in a sequence, over each person's sequences, and over people, so every person counts equally. Let $E_{i,\cdot}$ be the error of person $i$ under a method. We compare a rival with \name by $d=\operatorname{mean}_i\log(E_{i,\mathrm{rival}}/E_{i,\mathrm{FLINCH}})$ and report the reduction $100(1-e^{-d})\%$, the drop in squared error given by the geometric-mean ratio across people. Tables report person-mean RMS error in micrometres ($\mu$m), so their ratios differ from these reductions. Intervals are 95\% intervals from 10,000 resamples of people. Accuracy results use parameters before rounding to bits, so they measure the bases, not the rounding. \cref{tab:sealed} lists every compared method with references.

\subsection{Accuracy at a fixed code size}
\label{sec:main}
\begin{figure*}[t]
\centering
\includegraphics[width=\textwidth]{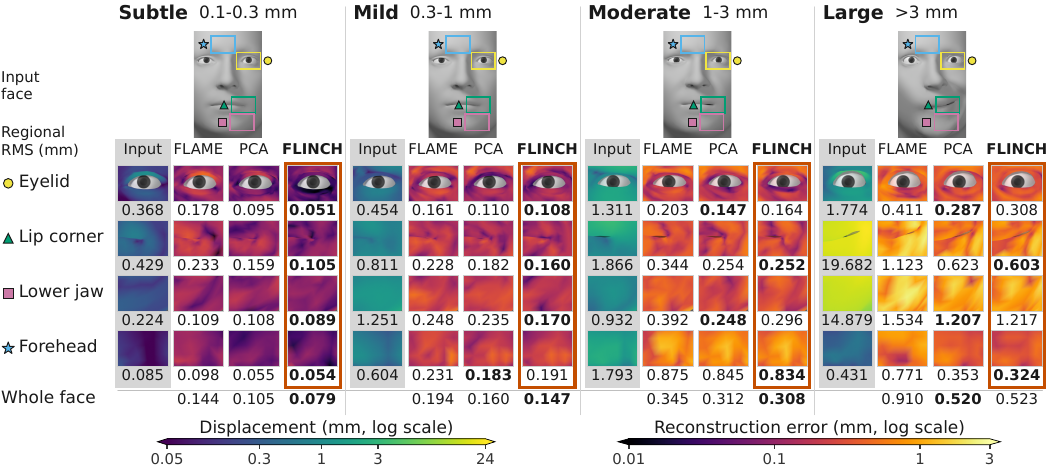}
\caption{\protect\input{figures/fig1grid_results_caption}}
\label{fig:grid}
\end{figure*}
\begin{figure}[t]
\centering
\includegraphics[width=\columnwidth,trim=0 4 0 4,clip]{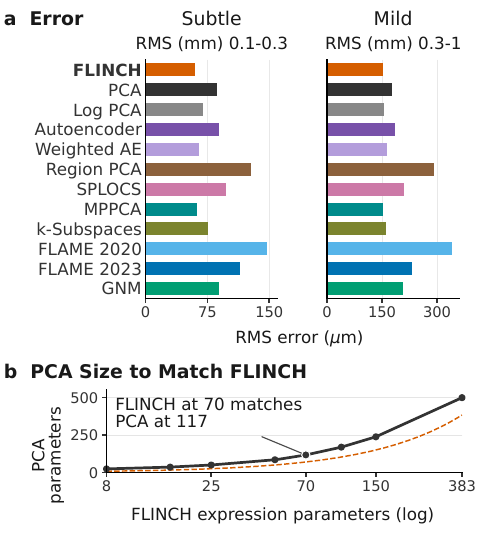}
\caption{\protect\input{figures/fig3_error_by_band_caption}}
\label{fig:bands}
\end{figure}

Does \name beat a single basis with the same number of parameters? At 70 parameters, \name has 53, 26 and 10\% lower squared error than PCA trained on the same data on Subtle, Mild and Moderate expressions (\cref{tab:sealed,fig:grid,fig:bands}). On Large expressions the reduction is 0.4\%, interval $[-0.6,1.3]$, so the gain on low-intensity expressions comes with no measurable loss on Large ones. \name has lower error than PCA on Subtle expressions at every code size tested, from 8 to 383 parameters. Above 70 parameters, its error on Large expressions is 2 to 9\% higher than PCA's. PCA needs 117 parameters, interval $[115,119]$, to match \name at 70 parameters on Subtle expressions (\cref{fig:bands}b). Log PCA, which gives small errors more weight, comes closer, but \name still has 26\% lower error on Subtle expressions. Against region PCA, the reductions from Subtle to Large are 78, 73, 66 and 59\%, and against mesh autoencoders 55, 32, 25 and 27\%. An autoencoder trained with intensity weights comes closest, and \name still has 15\% lower error on Subtle expressions.

\begin{table}[t]
\centering\sffamily
\setlength{\tabcolsep}{2pt}
\fontsize{\figlabelsize}{\figtableleading}\selectfont
\definecolor{blockblue}{RGB}{236,236,236}\definecolor{flinchtint}{RGB}{253,232,208}
\begin{tabular*}{\columnwidth}{@{\extracolsep{\fill}}lrrrr@{}}
\toprule
Method & \shortstack{Subtle\\0.1-0.3} & \shortstack{Mild\\0.3-1} & \shortstack{Moderate\\1-3} & \shortstack{Large\\$>3$} \\
\midrule
\rowcolor{blockblue} \multicolumn{5}{@{}l@{}}{\textit{Shared fits}} \\
\rowcolor{flinchtint} \textbf{FLINCH} & \textbf{59.5} & \textbf{151.6} & \textbf{268.3} & \textbf{472.1} \\
PCA & 86.4 & 176.6 & 282.0 & 473.4 \\
Log PCA & 69.5 & 155.5 & 272.3 & 509.2 \\
FLAME 2020 & 147.6 & 340.8 & 587.5 & 1176.1 \\
FLAME 2023 & 114.6 & 231.1 & 372.0 & 722.3 \\
FLAME 2023 (103) & 91.6 & 181.7 & 289.2 & 574.2 \\
GNM (383) & 88.9 & 207.2 & 398.6 & 1233.2 \\
OVR-style rig (72) & 173.5 & 467.6 & 952.0 & 2730.8 \\
\addlinespace[2pt]
\rowcolor{blockblue} \multicolumn{5}{@{}l@{}}{\textit{Five-fold neural and local models}} \\
\rowcolor{flinchtint} \textbf{FLINCH} & \textbf{59.5} & \textbf{151.6} & \textbf{268.1} & \textbf{473.7} \\
Autoencoder$^\dagger$ & 88.8 & 184.4 & 309.4 & 554.7 \\
Weighted Autoencoder$^\dagger$ & 64.5 & 163.4 & 315.1 & 660.4 \\
Region PCA & 128.4 & 292.7 & 460.8 & 738.3 \\
SPLOCS$^\ddagger$ & 97.4 & 209.5 & 343.1 & 573.9 \\
\addlinespace[2pt]
\rowcolor{blockblue} \multicolumn{5}{@{}l@{}}{\textit{Mixture models}} \\
\rowcolor{flinchtint} \textbf{FLINCH} & \textbf{58.9} & \textbf{149.6} & \textbf{265.0} & \textbf{468.6} \\
MPPCA & 62.9 & 151.5 & \textbf{265.0} & 469.4 \\
k-Subspaces & 75.6 & 161.2 & 270.4 & 473.3 \\
\addlinespace[2pt]
\bottomrule
\end{tabular*}

\caption{\protect\textbf{FLINCH has the lowest error on Subtle expressions in every comparison.} Entries are person-mean RMS reconstruction errors in $\mu$m on 20 held-out FaMoS people, one column per intensity band (lower is better). Methods use 70 expression parameters unless a size is in brackets. FLAME 2023~\cite{flame2017} appears with 67 expression components plus jaw, and in full with all 100 plus jaw (103). GNM~\cite{gnm2026} (383) and the OVR-style rig (72), built in FLAME topology, are used at full size, the rig fitted by least squares. The other rows are PCA, Log PCA (fitted to mean person log squared error), region PCA~\cite{tena2011}, SPLOCS~\cite{splocs2013}, mesh autoencoders~\cite{coma2018,gong2019spiralnet}, one weighted by intensity, mixtures of probabilistic PCA (MPPCA)~\cite{tipping1999mixtures} and k-Subspaces~\cite{vidal2011subspace}. Each block has its own FLINCH fit. Grey rows name blocks, orange rows mark FLINCH and bold marks the lowest error per block and column.
$^\dagger$Trained on 48 people plus 12 for early stopping, tested on all 20.
$^\ddagger$Did not converge. The supplement gives 95\% person intervals, and \cref{tab:generality} gives Ava-256 results.
}
\label{tab:sealed}
\end{table}

\paragraph{Published expression spaces.}
\label{sec:mixture}
Against FLAME 2023 at 70 parameters with a shared jaw, the reductions are 73, 57, 48 and 57\%. PCA is fitted on the same FaMoS training people, while FLAME is a published model learned from other 3D scan data. Even with all 100 expression components plus jaw (103 parameters), FLAME has higher error than PCA at 70 parameters in every band. \name at 70 parameters lowers its squared error by 58, 30, 14 and 32\%. FLAME 2020, the expression space used by DECA and EMOCA, has higher error than FLAME 2023 (\cref{tab:sealed}). GNM~\cite{gnm2026}, with 383 parameters, has higher error than \name at 70 parameters in every band. Against the 72-weight OVR-style VR headset rig in FLAME topology, \name lowers squared error by 88, 89, 92 and 97\%. The rig's error is highest on Large expressions, so its weakness is general (supplement).

\paragraph{Is \name just a mixture of subspaces?}
At 70 parameters, \name has lower error on Subtle expressions than mixture models, which learn subspaces without intensity: 12\% lower than MPPCA, 8\% lower than MPPCA fitted without means and 39\% lower than $k$-subspaces. Given intensity groups to start from, $k$-subspaces comes within 3\% (supplement). Intensity is therefore an effective grouping, and \name builds it in directly, without fitting a mixture.

\subsection{Refitting and ablations}
\label{sec:transfer}
\begin{table}[t]
\centering\sffamily
\setlength{\tabcolsep}{2pt}
\fontsize{\figlabelsize}{\figtableleading}\selectfont
\definecolor{blockblue}{RGB}{236,236,236}\definecolor{flinchtint}{RGB}{253,232,208}
\begin{tabular*}{\columnwidth}{@{\extracolsep{\fill}}lrrrr@{}}
\toprule
Method & \shortstack{Subtle\\0.1-0.3} & \shortstack{Mild\\0.3-1} & \shortstack{Moderate\\1-3} & \shortstack{Large\\$>3$} \\
\midrule
\rowcolor{blockblue} \multicolumn{5}{@{}l@{}}{\textit{Ava-256 (estimated $\mu$m)}} \\
\rowcolor{flinchtint} \textbf{FLINCH} & 74.5 & \textbf{262.5} & \textbf{515.4} & \textbf{894.6} \\
PCA & 81.5 & 283.2 & 552.6 & 918.7 \\
Log PCA & \textbf{73.8} & 263.4 & 531.0 & 949.1 \\
FLAME Open & 102.6 & 405.5 & 914.9 & 1816.7 \\
\textbf{FLINCH (FaMoS)} & 91.1 & 346.2 & 769.1 & 1518.7 \\
\addlinespace[2pt]
\bottomrule
\end{tabular*}

\caption{\protect\textbf{Refitted on Ava-256, FLINCH has lower error than PCA in every intensity band.} Entries are person-mean RMS reconstruction errors in $\mu$m on an estimated metric scale, one column per intensity band, for 176 Ava-256 people, each scored by models fitted without them (lower is better). Grey rows head each block, the orange row highlights FLINCH and bold marks the lowest rounded error per block and column, including ties. The supplement gives 95\% person intervals and FaMoS ablations.\par {\footnotesize Abbreviations as in \cref{tab:sealed}. FLAME Open is the open release of FLAME 2023.}
}
\label{tab:generality}
\end{table}

\begin{figure*}[t]
\centering
\includegraphics[width=\textwidth,trim=0 0 0 6,clip]{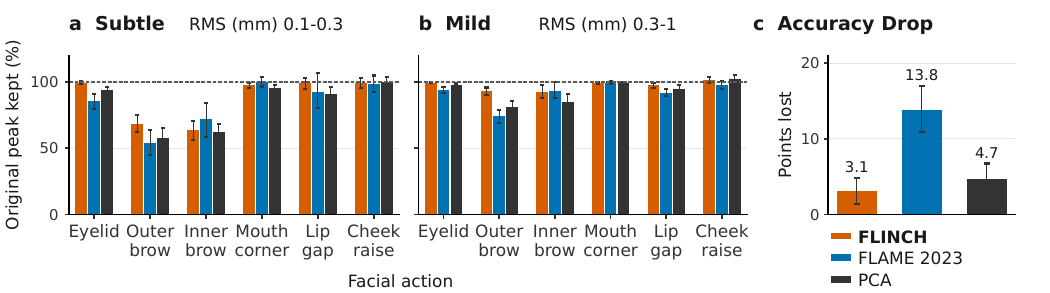}
\caption{\protect\input{figures/fig4_fidelity_caption}}
\label{fig:exprrec}
\label{fig:fidelity}
\end{figure*}
Does \name carry over to another capture setup? Refitted on the Ava-256 dataset~\cite{ava2562024}, \name has 16, 15, 14 and 6\% lower error than PCA from Subtle to Large (\cref{tab:generality}). In this cross-validation over all 213 Ava-256 people, scored on the 176 with frames in every band, each person is scored by a fit that excludes them, and Ava-256's official held-out people are never used. Log PCA has 2\% lower squared error than \name on Subtle expressions here, and \name has lower error than Log PCA on Moderate and Large expressions. FaMoS-trained libraries, like any learned basis, are best refitted for a new capture setup.
Even without refitting, they have lower error than FLAME Open in every band (\cref{tab:generality}).
\label{sec:ablation}
In FaMoS ablations (supplement), four bands lower error on Subtle expressions by 20\% compared with two. More than four bands bring only small further gains and raise error on Large expressions. Choosing the library by fit, rather than by measured band, lowers error on Mild expressions by 9\%. Over time, \name switches library only when another one fits better by a fixed margin. This cuts library switches on Mild Ava-256 test frames from 8.1\% to 0.02\%, and frame-to-frame error falls from 5\% above PCA's to 6\% below.

\begin{figure*}[t]
\centering
\includegraphics[width=\textwidth,trim=0 0 0 3,clip]{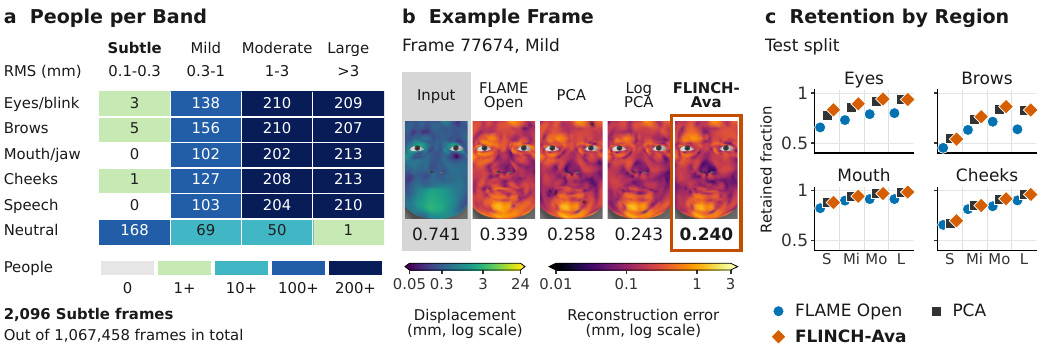}
\caption{\protect\input{figures/fig6_dataset_caption}}
\label{fig:dataset}
\end{figure*}

\subsection{Beyond average error}
\label{sec:exprrec}
\paragraph{Facial actions.}
Peak retention is the reconstructed peak of a facial action divided by its original peak, so 100\% keeps the peak and larger values overshoot. On Subtle expressions, \name keeps 99\% of the eyelid peak, against 94\% for PCA and 86\% for FLAME 2023 (\cref{fig:fidelity}). On Ava-256, \name keeps 86\% of Subtle eye-region peaks over 98 expression events from 89 people, against 80\% for PCA and 65\% for FLAME 2023, and Log PCA keeps a similar share (supplement). A classifier trained on the original meshes reaches 43.6\% accuracy. It loses 3.1 points on \name reconstructions and 13.8 points on FLAME 2023 reconstructions, and the supplement breaks this down by band.

\paragraph{Bits per frame.}
\label{sec:bitrate}
With each parameter rounded to a fixed number of bits, PCA needs 910 bits per frame (91 parameters at 10 bits) to match the error of \name at 562 bits on Subtle expressions, so \name uses 38\% fewer bits. Giving each library its own number of parameters saves a further 9 to 14\% (supplement).

\paragraph{Mixed expressions.}
\label{sec:mixed}
When the lower-face expression is much larger than the upper-face expression, the encoder may favour the library that fits the upper face. At the same 562 bits, this cuts upper-face error by 30\%, while whole-face error rises by 0.02\% (supplement).

\paragraph{Real headset signals and image input.}
\label{sec:real}
We drive the OVR-style VR headset rig with 76 recorded micro-expression events of 20 people from BlendME~\cite{blendme2026}. \name keeps 1.2 percentage points more of each event's peak than PCA, 95\% interval $[0.3,2.5]$, and matches FLAME (supplement). A tracker trained on rendered Ava-256 meshes has one output head each for PCA, \name and FLAME parameters. On 20 held-out people, it drives \name within 0.3\% of PCA's error in every band, and below FLAME's error in all four (supplement).

\section{FLINCH-4D expression benchmark}
\label{sec:data}
FLINCH-4D measures how often each intensity band occurs in recorded faces and how well expression representations preserve each band. It labels 1,067,458 registered Ava-256 frames~\cite{ava2562024} of 213 people. Ava-256 records each person in short scripted clips, such as blinks, brow, jaw and cheek movements, speech and a held neutral face, 11,407 clips in total. Displacement is measured from each person's neutral frame after head pose is removed, in estimated millimetres because Ava-256 has no calibrated metric scale. Each frame is labelled with its whole-face displacement, its intensity band and the displacement of six facial regions: eyes, brows, mouth, cheeks, forehead and nose. Regions are fixed vertex masks, drawn before any evaluation. Each expression event (\cref{sec:code}) is labelled with its start, peak and end frames. \Cref{fig:dataset} shows, for six of these clip types, how many people have frames in each band, and how much of each region's displacement each method preserves. The supplement compares FLINCH-4D with other 4D expression datasets.

\paragraph{Splits and results.}
The 213 people are divided 129/42/42 into training, validation and test splits. Test-split people are never used for training or model selection. At 70 parameters, FLINCH-Ava, which is FLINCH fitted on the training split, has the lowest whole-face RMS error on Mild, Moderate and Large expressions: 0.270/0.531/1.110~mm, against 0.290/0.568/1.126 for PCA, 0.275/0.554/1.229 for Log PCA and 0.401/0.926/2.333 for FLAME Open. On Subtle expressions, Log PCA is lowest at 0.070~mm, with FLINCH-Ava at 0.071 and PCA at 0.076. Errors are person means, as in \cref{tab:sealed}, and the supplement adds the autoencoder. Table~\ref{tab:generality} uses a separate cross-validation over all 213 people, scored on the 176 with frames in every band, so it includes this test split and its values differ. Ava-256's scripted clips contain few Subtle expressions: only 2,096 frames, most of them in the neutral-face clips, where faces move least (\cref{fig:dataset}a).

\paragraph{Spontaneous micro-expressions in 4D.}
\label{sec:spontaneous}
4DME~\cite{fourDME2023} records spontaneous micro-expressions as real 4D face scans, 3D face geometry over time, so it tests \name directly on spontaneous expressions. On all 22 analysable 4DME clips from 12 people, with the libraries unchanged, \name's RMS error over the face area of each annotated facial action is 2\% lower than PCA's and 9\% lower than FLAME 2023's at 70 parameters, with person-level 95\% intervals above zero for both. It is lower than FLAME's for all 12 people.

\paragraph{Release.}
Labels, splits, region masks, evaluation code, FLINCH-Ava libraries, per-frame displacements and scripts that rebuild them from the original Ava-256 data will be released. Raw Ava-256 meshes, images and textures are not redistributed, and no data derived from FaMoS is included.

\section{Conclusion and future work}
\label{sec:conclusion}
\paragraph{Scope and future work.}
\name selects its library from the 3D displacement of each frame, so it applies directly in capture and animation pipelines, and image or headset input needs a tracker that predicts its code. A tracker trained on rendered images already drives \name within 0.3\% of PCA's error (\cref{sec:real}), and we are extending this to real headset cameras. Our 4D evaluation combines posed and scripted scans, spontaneous headset recordings and a first test on spontaneous 4D micro-expression scans (\cref{sec:data}), which we are extending to more people. Our measures are geometric, and a perceptual study of how viewers judge \name avatars is planned. FLINCH-4D labels describe displacement rather than facial physiology. The supplement gives full protocol details.

\paragraph{Conclusion.}
We have presented \name, an intensity-conditioned expression representation that gives each intensity band its own library. On FaMoS at 70 parameters, it preserves low-intensity (Subtle and Mild) expressions that a single basis of the same size loses, with no measurable loss on Large ones and only 2 extra bits per frame. We have shown that evaluating models per intensity band exposes losses that a single average error hides, and we have introduced FLINCH-4D, a 4D benchmark labelled by expression intensity, to support such evaluation at scale. With linear decoding and a compact code, \name can strengthen pipelines that send expression parameters, such as those for headset telepresence and film. Preserving low-intensity expressions, such as a faint smile or a moment of hesitation, brings avatars closer to the realism people expect from faces. Future work includes learned bands and intensity conditioning for neural and Gaussian avatars, toward face models that preserve the whole range of expression.

\section*{LLM Usage Statement}
Anthropic coding agents were used to implement and debug parts of the training and evaluation pipelines and assist with text editing. The authors designed the study, verified all results and citations, and take full responsibility for the submitted work.

\section*{Acknowledgements}
This work was supported by the Dolby Laboratories University Research Collaboration program.

{
  \small
  \linespread{0.97}\selectfont
  \setlength{\bibsep}{0pt}
  \bibliographystyle{flinch-references}
  \bibliography{references}
}

\clearpage
\appendix
\section*{Appendix: Supplementary Material}
\setcounter{figure}{0}
\setcounter{table}{0}
\setcounter{equation}{0}
\renewcommand{\thefigure}{S\arabic{figure}}
\renewcommand{\thetable}{S\arabic{table}}
\renewcommand{\theequation}{S\arabic{equation}}
\renewcommand{\theHfigure}{S\arabic{figure}}
\renewcommand{\theHtable}{S\arabic{table}}
\renewcommand{\theHequation}{S\arabic{equation}}
\definecolor{flinchtint}{HTML}{FDE8D0}
\definecolor{blockgrey}{HTML}{ECECEC}
\setlength{\textfloatsep}{8pt plus 2pt minus 2pt}
\setlength{\dbltextfloatsep}{8pt plus 2pt minus 2pt}
\setlength{\floatsep}{6pt plus 2pt minus 2pt}
\setlength{\dblfloatsep}{6pt plus 2pt minus 2pt}
\captionsetup{font=small,skip=3pt}
\makeatletter
\renewcommand\paragraph{\@startsection{paragraph}{4}{\z@}
  {.75ex plus .15ex minus .05ex}{-1em}{\normalfont\normalsize\bfseries}}
\makeatother
\section{Overview}
\label{sec:s-overview}
\name (Facial Libraries for INtensity-Conditioned Human expressions) is an intensity-conditioned expression representation. It has one shape library per intensity band: Subtle, Mild, Moderate and Large. Each frame is stored as the parameters of whichever library reconstructs it best, plus a 2-bit index. On FaMoS at 70 parameters, \name preserves low-intensity (Subtle and Mild) expressions that a single basis of the same size loses, with no measurable loss on Large ones (main-paper Sec.~\ref{sec:main}).

\Cref{sec:s-why} gives method and theory details, \cref{sec:s-protocol} the implementation and evaluation protocol, and \cref{sec:s-card} the datasets. \Cref{sec:s-results,sec:s-meaning} add reconstruction and expression-preservation results, \cref{sec:s-inputs} covers real headset recordings, rendered images and spontaneous 4D scans, and \cref{sec:s-ethics} covers ethics and release.
\begin{figure*}[!tp]
\centering
\includegraphics[width=\textwidth]{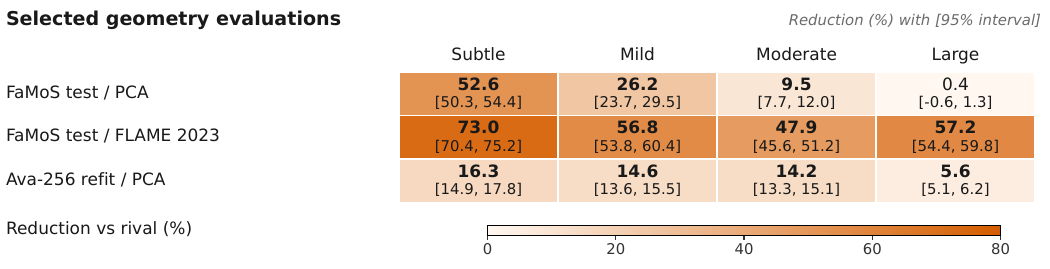}
\caption{\textbf{FLINCH has lower squared error than PCA on Subtle to Moderate expressions.} Each row compares 70-parameter geometry fits with the named rival. FaMoS rows use 20 test people. The Ava-256 row is refitted without each scored person and includes 176 people with frames in all four bands. Each cell gives how much lower FLINCH's squared error is than the rival's, in percent, with a 95\% interval in brackets. For example, 52.6 [50.3, 54.4] in the Subtle column of the FaMoS test / PCA row means FLINCH has 52.6\% lower squared error than PCA on Subtle expressions, and the 95\% interval over people runs from 50.3 to 54.4\%. Positive values favour FLINCH. Darker orange means a larger reduction. Each person counts equally, and intervals come from resamples of people with the models unchanged. PCA means principal component analysis. FLAME 2023 is the released face model. Bands use whole-face root-mean-square displacement from neutral: Subtle 0.1--0.3, Mild 0.3--1, Moderate 1--3, Large $>3$\,mm. Bold marks reductions whose 95\% interval is above zero, that is, a clear gain for FLINCH.}
\label{fig:s-evidence}
\end{figure*}

\paragraph{Evaluation data.}
\label{sec:s-limits}
Evaluation covers posed and scripted 3D scans, spontaneous headset recordings and spontaneous 4D micro-expression scans.

\section{Method and theory details}
\label{sec:s-why}
\paragraph{Why one basis loses Subtle expressions.}
\label{sec:s-theory}
A single basis, such as PCA fitted to all frames together, is the standard expression code that \name replaces. At $K=70$ parameters, both use 67 expression parameters and 3 jaw parameters (main-paper Sec.~\ref{sec:encoding}). The analysis below concerns the 67 expression directions. On Subtle training expressions, the single basis loses 15.4\% of the expression energy. The Subtle library of \name, with the same 67 directions, loses 5.6\%. Main-paper Sec.~\ref{sec:one-basis} states that a perturbation bound accounts for 81 to 83\% of this excess across folds. We give the bound here in two parts.

First, as in main-paper Sec.~\ref{sec:one-basis}, let $S_b$ be the average of $zz^\top$ over band $b$'s training frames, where $z$ is the expression with its jaw part removed. Its eigenvectors $u_{b,i}$ are the directions that the band's expressions occupy, and its eigenvalues $\lambda_{b,i}$ measure how strongly. Let $T_b=\operatorname{tr}S_b$ be the band's total energy, $\tau_b$ the share that the band's own best $r=67$ directions lose, $P$ the projection onto the single basis, and $\ell_i=\lVert(I-P)u_{b,i}\rVert^2$ the part of direction $u_{b,i}$ that the single basis loses. Then
\begin{equation}
\frac{\operatorname{tr}((I-P)S_b)}{T_b}\geq\tau_b+\frac{\sum_{i\leq r}(\lambda_{b,i}-\lambda_{b,r+1})\ell_i}{T_b}.
\label{eq:s-bound}
\end{equation}
The sum adds up the band's main directions that the single basis loses, weighted by their strength. This bound accounts for 81 to 83\% of the excess.

Second, let $A$ be the same average over the other bands, weighted by their shares, $Q$ the projection onto its best 67 directions, and $\delta>0$ the gap between its 67th and 68th eigenvalues. Adding band $b$ with share $\pi_b$ moves the directions by at most $\lVert P-Q\rVert_F/\sqrt{2}\leq\sqrt{2}\lVert\pi_bS_b\rVert_F/\delta$. Subtle expressions carry little energy, so the right side is small, and the single basis stays close to the directions of larger expressions. The bound describes the training fits. \Cref{sec:s-results,sec:s-meaning} report results on held-out people.

\paragraph{Are small expressions scaled-down large ones?}
\label{sec:s-shape}
An expression event is a run of consecutive non-neutral frames (main-paper Sec.~\ref{sec:code}). Its peak frame is the frame with the largest displacement. If small expressions were scaled-down large ones, one basis would be enough (main-paper Sec.~\ref{sec:one-basis}). We test this on the frames before each event's peak. For each such frame in the Mild band, we scale the peak frame to match it as closely as possible and measure what is left. On average, $\rho=0.909$, 95\% interval $[0.903, 0.915]$, of the frame's RMS displacement is left. Capture noise alone, measured on pairs of neutral frames of the same person, leaves 0.667 $[0.637, 0.695]$. With slow changes over time removed first, noise alone leaves 0.937 $[0.926, 0.947]$. The test needs at least 10 people with at least 20 such frames in a band. In the Mild band, 150 Ava-256 people meet this requirement. The reconstruction comparisons in \cref{sec:s-results,sec:s-meaning} test the benefit of separate libraries directly, on held-out people.

\section{Implementation and evaluation protocol}
\label{sec:s-protocol}
\label{sec:s-bands}
\paragraph{Implementation details.}
\Cref{tab:s-implementation,tab:s-baselines} give the full protocol details of FLINCH and of every compared method. Intensity is the root-mean-square (RMS) displacement of the face from its neutral shape after head pose is removed. Accuracy results use parameters before rounding to bits. Error is the squared distance between reconstructed and true vertices, averaged over vertices. \Cref{tab:s-ablate} compares the rounded band limits with exactly log-spaced ones.

\begin{table*}[!tp]

\centering\sffamily\fontsize{\figlabelsize}{\figtableleading}\selectfont
\setlength{\tabcolsep}{4pt}\renewcommand{\arraystretch}{1}
\begin{tabularx}{\textwidth}{@{}L{0.20\textwidth}Y@{}}
\toprule
Item & Details \\
\midrule
Intensity bands (mm) & Subtle $[0.1, 0.3)$, Mild $[0.3, 1)$, Moderate $[1, 3)$, Large $>3$. Frames below 0.1\,mm count as neutral and are not fitted or scored. \\
Code and neutral face & $K=70$ parameters: 67 expression parameters plus 3 jaw parameters shared by all libraries. Libraries have no mean, so all-zero parameters give the neutral face. \\
Training frames & Every Subtle frame and at most 128 frames from the other bands per person. Every learned method uses the same frames. \\
Library choice & Each frame uses the library with the lowest whole-face fit error, chosen frame by frame. \\
Optional switching over time & Switches library only when another one fits better by a fixed margin of 0.066 estimated mm$^2$ per vertex, the 75th percentile of the positive switching gains in training. \\
Optional mixed-expression coding & A frame is mixed when one of the eyes, brows, mouth and cheeks moves less than 0.3 and another more than 1 estimated mm. Candidate region weights are 1, 0.5 and 0.25. Frames that are not mixed keep their default code. Each recording's total whole-face squared error stays within 1\% of the default. Each recording is optimised within 30\,s. \\
Bits & 8 bits per parameter plus a 2-bit library index, 562 bits at $K=70$. Each library's rounding ranges are set on its training frames. \\
Seeds: frames and mixtures & Frame sampling: $20261031+$ numeric person identifier. Mixture fits: $5000+K$. \\
Seeds: neural models & Autoencoder: $20261110+100f+K$, fold $f=0,\ldots,4$. Image tracker output heads: 20261112, 20261113, 20261114. \\
Intervals & 95\% intervals from 10,000 resamples of people, seed 20261028 unless stated. 4DME uses 20261114. \\
\bottomrule
\end{tabularx}
\caption{\textbf{FLINCH uses the same number of parameters as a single basis, plus a 2-bit library index.} Rows give the protocol details of the four libraries, their fitting and the evaluation. $K$ is the number of parameters per frame, including the jaw. RMS means root-mean-square displacement from neutral, and mm means millimetres. Ava-256 distances are in estimated mm because Ava-256 has no calibrated metric scale. The two optional variants are not used unless stated. Seeds make random choices repeatable, and $f$ is the fold index.}
\label{tab:s-implementation}
\end{table*}
\begin{table*}[!tp]
\centering\sffamily\fontsize{\figlabelsize}{\figtableleading}\selectfont
\setlength{\tabcolsep}{3pt}\renewcommand{\arraystretch}{1}
\begin{tabularx}{\textwidth}{@{}L{0.14\textwidth}L{0.14\textwidth}L{0.095\textwidth}YL{0.12\textwidth}@{}}
\toprule
Method & Training data & Parameters & Fitting & Reference \\
\midrule
\rowcolor{blockgrey}\multicolumn{5}{@{}l}{\textit{Shared fits}}\\
PCA & FaMoS, 60 per fold & 67+3 & PCA without a mean, fitted after removing the jaw. Exact projection onto its directions. & \cite{eckart1936approximation} \\
Log PCA & Same frames & 67+3 & Minimises the mean log squared error over people and bands, floor $10^{-12}$. Ordinary and weighted starts. At most 200 weighted PCA updates, relative tolerance $10^{-7}$. & main-paper Sec.~\ref{sec:main} \\
FLAME 2023 and FLAME 2020 & Released bases & 70, and 103 for FLAME 2023 & Released expression directions plus the shared jaw, fitted by exact least squares. FLAME's pose-dependent shape corrections are not used. & \cite{flame2017} \\
GNM & Released basis & 383 & All released directions, mapped to the FaMoS face, fitted by exact least squares. Used as released, without refitting. & \cite{gnm2026} \\
OVR-style rig & Fixed BlendME rig & 72 & Least squares with weights of any value or weights in $[0, 1]$. Tolerance $10^{-10}$, at most 1,000 iterations, optimality check below $10^{-7}$. & \cite{blendme2026} \\
\rowcolor{blockgrey}\multicolumn{5}{@{}l}{\textit{Five-fold neural and local models}}\\
Autoencoders & 48 fitting and 12 stopping people per fold & 67+3 & SpiralNet++ mesh autoencoder with spiral length 9, channels 32/32/32/64 and 4 pooling steps of factor 4. Adam, rate $10^{-3}$, decay 0.99, batch 32, at most 300 epochs, early stopping after 30. Ordinary losses and losses weighted by intensity. & \cite{gong2019spiralnet} \\
Region PCA & FaMoS, 60 per fold & 67+3 & PCA in three separate face regions, directions ranked by training energy, then projected jointly. & \cite{tena2011} \\
SPLOCS$^\ddagger$ & FaMoS, 60 per fold & 67+3 & 67 sparse local components, $\lambda=0.01$. At most 100 outer and 500 inner iterations. Scored by projection onto these components. & \cite{splocs2013} \\
\rowcolor{blockgrey}\multicolumn{5}{@{}l}{\textit{Mixture models}}\\
MPPCA & FaMoS, 75 non-test people & 4 subspaces & Each subspace has 67 expression directions after jaw removal, plus the jaw. 5 weighted expectation--maximisation starts, at most 80 iterations, tolerance $10^{-5}$. The chosen subspace sends 67+3 parameters plus a 2-bit index. Subspace means are not decoded. & \cite{tipping1999mixtures} \\
$k$-subspaces & FaMoS, 75 non-test people & 4 subspaces & Two weighted versions, one with 5 random starts and one started from the intensity bands. At most 100 rounds of reassigning frames and refitting PCA, tolerance $10^{-6}$. & \cite{vidal2011subspace} \\
\rowcolor{blockgrey}\multicolumn{5}{@{}l}{\textit{Image prediction}}\\
Image tracker & Ava-256, 56/18/20 & 70 per output head & One shared image encoder for all codes. AdamW, 8,000 encoder and 6,000 output head updates per seed, head rate $10^{-4}$, batch 512. Libraries are combined by weights over 271 independent directions. & main-paper Sec.~\ref{sec:exprrec} \\
\bottomrule
\end{tabularx}
\caption{\textbf{Every expression code is evaluated on the same face geometry.} Rows specify fitting data, number of parameters and fitting method. Sizes include the jaw. FaMoS is the 3D scan dataset, with 20 test people. The image tracker uses rendered images of 56, 18 and 20 Ava-256 people for training, validation and testing. PCA is principal component analysis. Log PCA gives small errors more weight during fitting. FLAME and GNM are released face models. SPLOCS learns sparse local deformations. MPPCA is a mixture of probabilistic PCA models. The OVR-style rig is a fixed headset rig. The autoencoders learn nonlinear mesh codes. Each block of rows has its own matched FLINCH fit. The last column cites each method. $^\ddagger$SPLOCS did not converge.}
\label{tab:s-baselines}
\end{table*}
\paragraph{Data and protocol.}
We score every model on the same 1,787 face vertices, with each person weighted equally. FaMoS has 95 people: 75 in 5 folds of 60 fitting and 15 validation people, and 20 separate test people. The first frame of each sequence is neutral. Ava-256 refitting uses 5 folds over 213 people. Each person is scored by a fit that excludes them, and the 176 people with frames in all four bands are scored. Ava-256's own held-out people, separate from these 213, are never used (main-paper Sec.~\ref{sec:transfer}). FLINCH-4D splits the 213 people into 129 for training, 42 for validation and 42 for testing. As noted in main-paper Sec.~\ref{sec:data}, the cross-validation over all 213 people (\cref{tab:s-tab2}) includes the FLINCH-4D test people, and each scored person is excluded from their own fit. Test-split people are never used for training or model selection of the benchmark models in \cref{tab:s-natural}. Ava-256 has no calibrated metric scale, so its distances are in estimated millimetres. One scale is used for the whole dataset, set by an assumed 63\,mm distance between the eyes.

The three jaw directions are small rotations of the jaw about its neutral pivot, weighted by the skinning weights and made orthonormal. Their part is removed from the training frames before each library is fitted. Parameters can be interpolated within a library, and each library has its own parameter coordinates. Squared error is averaged over the frames of each band in a sequence, then over each person's sequences, then over people, so every person counts equally (main-paper Sec.~\ref{sec:setup}). A person counts for a band if they have at least five frames in it. For person errors $E_i$, the reduction is $100(1-e^{-d})\%$, where $d=\operatorname{mean}_i\log(E_{i,\mathrm{rival}}/E_{i,\mathrm{FLINCH}})$. Tables report person-mean RMS error, so their ratios differ from these reductions. The 95\% intervals resample people with the fitted models unchanged. They use bootstrap-$t$ for PCA, FLAME, local models and autoencoders, and percentiles for mixtures and downstream tasks.

\paragraph{Names and codes.}
FLAME Open is the open release of FLAME 2023 (main-paper Tab.~\ref{tab:generality}). Sizes include the jaw. FLINCH-Ava is FLINCH fitted on the FLINCH-4D training split (main-paper Sec.~\ref{sec:data}).

\paragraph{How common are low-intensity expressions in headset recordings?}
\label{sec:s-census}
Low-intensity expressions are common in real use (main-paper Sec.~\ref{sec:intro}). In BlendME~\cite{blendme2026}, spontaneous headset recordings of people suppressing their expressions, 69\% of the frames that show an expression show a low-intensity one, 95\% interval $[59, 79]$. In EmojiHeroVR~\cite{ortmann2024emojiherovr}, where people pose expressions on cue in a virtual reality game, the share is 22\% $[18, 26]$. Both datasets record the headset tracker's named weights, which drive the same fixed 72-weight OVR-style rig. All-zero weights give the neutral face. Each frame's band is set by its RMS displacement over the 1,787 face vertices, in mm. Frames at or above 0.1\,mm count as showing an expression, and shares are averaged equally over people.

\section{Datasets and measurement}
\label{sec:s-card}
FLINCH-4D labels 1,067,458 registered Ava-256 frames of 213 people, in 11,407 clips with 14,792 expression events. Each frame is labelled with its whole-face displacement, its intensity band and the displacement of six facial regions: eyes, brows, mouth, cheeks, forehead and nose. Each expression event is labelled with its start, peak and end frames. Frames also carry one of 26 scripted actions and mixed-expression and quality flags. People are split 129, 42 and 42 into training, validation and test sets, and no person appears in more than one set. A frame is mixed when one of the eyes, brows, mouth and cheeks moves less than 0.3 and another more than 1 estimated mm. This holds for 23,119 frames (2.2\%). As in main-paper Sec.~\ref{sec:data}, the scripted clips contain 2,096 Subtle frames, 1,686 (80.4\%) of them in the neutral-face clips. As stated in main-paper Sec.~\ref{sec:conclusion}, the labels describe displacement. \Cref{sec:s-4dme} evaluates spontaneous micro-expressions on 4DME scans. \Cref{tab:s-datasets} compares FLINCH-4D with other 4D face datasets.

\begin{table*}[!tp]
\centering\sffamily\fontsize{\figlabelsize}{\figtableleading}\selectfont
\begin{tabularx}{\textwidth}{@{}L{2.6cm}rL{2.0cm}L{3.2cm}L{1.6cm}Y@{}}
\toprule
Dataset & People & Clips & Frames & Frames per second & Labels \\
\midrule
CoMA~\cite{coma2018} & 12 & -- & 20,466 & -- & -- \\
D3DFACS~\cite{cosker2011d3dfacs} & 10 & 519 & -- & -- & peak facial-action coding \\
FaMoS~\cite{famos2023} & 95 & 28 per person & about 600,000 & 60 & -- \\
BU-4DFE~\cite{yin2008bu4dfe} & 101 & 606 & about 60,600 & 25 & six posed expressions \\
Multiface~\cite{wuu2022multiface} & 13 & -- & 12,200 to 23,000 per person & 30 & -- \\
Ava-256~\cite{ava2562024} & 256 & -- & about 5,000 or 10,000 per person & 7.5 or 15 & no intensity labels \\
\rowcolor{flinchtint}\textbf{FLINCH-4D} & \textbf{213} & \textbf{11,407} & \textbf{1,067,458} & 7.5 or 15 & \textbf{per-frame intensity band, RMS of 6 facial regions, 26 scripted actions} \\
\bottomrule
\end{tabularx}
\caption{\textbf{FLINCH-4D adds per-frame intensity labels to registered Ava-256 sequences.} Rows compare dataset release counts, capture rates in frames per second and available labels. Orange and bold highlight FLINCH-4D. 4D means 3D face geometry over time. RMS means root-mean-square displacement from neutral. FLINCH-4D keeps the Ava-256 frame rate and adds whole-face bands, displacement of six facial regions and expression events. Dataset names identify the cited releases. A dash means the release does not state the value.}
\label{tab:s-datasets}
\end{table*}

\section{Additional reconstruction results}
\label{sec:s-results}
At 70 parameters, \name has 52.6\% lower squared error than PCA on Subtle expressions, with a 95\% interval $[50.3, 54.4]$. On Large expressions the reduction is 0.4\%, interval $[-0.6, 1.3]$, so the gain on low-intensity expressions comes with no measurable loss on Large ones (main-paper Sec.~\ref{sec:main}). \Cref{tab:s-main-rms} gives the RMS errors of main-paper Tab.~\ref{tab:sealed}, and \cref{tab:s-tab1} gives the paired reductions with their intervals. \name has lower error than PCA on Subtle expressions at every code size tested, from 8 to 383 parameters. Above 70 parameters, FLINCH keeps its lead on Subtle expressions, while PCA's error on Large expressions is 2 to 9\% lower, as in main-paper Sec.~\ref{sec:main}. PCA needs 117 parameters, interval $[115, 119]$, to match \name at 70 parameters on Subtle expressions.

\begin{table}[!htbp]
\centering\sffamily\fontsize{\figlabelsize}{\figtableleading}\selectfont
\setlength{\tabcolsep}{2pt}
\begin{tabularx}{\columnwidth}{@{}Yrrrr@{}}
\toprule
Method & \shortstack{Subtle\\0.1--0.3} & \shortstack{Mild\\0.3--1} & \shortstack{Moderate\\1--3} & \shortstack{Large\\$>3$} \\
\midrule
\rowcolor{blockgrey}\multicolumn{5}{@{}l}{\textit{Shared fits}} \\
\rowcolor{flinchtint}\textbf{\name} & \textbf{59.5} & \textbf{151.6} & \textbf{268.3} & \textbf{472.1} \\
PCA & 86.4 & 176.6 & 282.0 & 473.4 \\
Log PCA & 69.5 & 155.5 & 272.3 & 509.2 \\
FLAME 2020 & 147.6 & 340.8 & 587.5 & 1176.1 \\
FLAME 2023 & 114.6 & 231.1 & 372.0 & 722.3 \\
FLAME 2023 (103) & 91.6 & 181.7 & 289.2 & 574.2 \\
GNM (383) & 88.9 & 207.2 & 398.6 & 1233.2 \\
OVR-style rig (72) & 173.5 & 467.6 & 952.0 & 2730.8 \\
\addlinespace[2pt]
\rowcolor{blockgrey}\multicolumn{5}{@{}l}{\textit{Five-fold neural and local models}} \\
\rowcolor{flinchtint}\textbf{\name} & \textbf{59.5} & \textbf{151.6} & \textbf{268.1} & \textbf{473.7} \\
Autoencoder$^\dagger$ & 88.8 & 184.4 & 309.4 & 554.7 \\
Weighted Autoencoder$^\dagger$ & 64.5 & 163.4 & 315.1 & 660.4 \\
Region PCA & 128.4 & 292.7 & 460.8 & 738.3 \\
SPLOCS$^\ddagger$ & 97.4 & 209.5 & 343.1 & 573.9 \\
\addlinespace[2pt]
\rowcolor{blockgrey}\multicolumn{5}{@{}l}{\textit{Mixture models}} \\
\rowcolor{flinchtint}\textbf{\name} & \textbf{58.9} & \textbf{149.6} & \textbf{265.0} & \textbf{468.6} \\
MPPCA & 62.9 & 151.5 & \textbf{265.0} & 469.4 \\
$k$-subspaces & 75.6 & 161.2 & 270.4 & 473.3 \\
\addlinespace[2pt]
\bottomrule
\end{tabularx}
\caption{\textbf{FLINCH has the lowest error on Subtle expressions in every comparison.} Entries are root-mean-square reconstruction errors in micrometres ($\mu$m), averaged equally over 20 FaMoS test people. Lower is better. Band limits in the headers are whole-face RMS displacement from neutral, in mm. Methods use 70 parameters including the jaw unless shown in brackets. Grey rows head each block, orange rows highlight FLINCH and bold marks the lowest error per block and column. PCA is principal component analysis. Log PCA gives small errors more weight during fitting. FLAME and GNM are released face models. SPLOCS learns sparse local deformations. MPPCA is a mixture of probabilistic PCA models. The OVR-style rig is a fixed headset rig. $^\dagger$Trained on 48 people plus 12 for early stopping. $^\ddagger$SPLOCS did not converge. The $k$-subspaces row starts from random groups of frames.}
\label{tab:s-main-rms}
\end{table}
\begin{table*}[!tp]
\centering\sffamily\fontsize{\figlabelsize}{\figtableleading}\selectfont
\setlength{\tabcolsep}{5pt}
\begin{tabularx}{\textwidth}{@{}Yrrrr@{}}
\toprule
Method & Subtle 0.1--0.3 & Mild 0.3--1 & Moderate 1--3 & Large $>3$ \\
\midrule
\rowcolor{blockgrey}\multicolumn{5}{@{}l}{\textit{Shared fits}}\\
PCA & 52.6 [50.3, 54.4] & 26.2 [23.7, 29.5] & 9.5 [7.7, 12.0] & 0.4 [$-0.6$, 1.3] \\
Log PCA & 26.5 [24.9, 28.2] & 4.8 [3.5, 6.3] & 2.9 [$-0.7$, 4.3] & 13.7 [11.2, 15.8] \\
FLAME 2020 & 83.8 [82.3, 85.0] & 80.1 [77.9, 82.2] & 79.2 [77.2, 80.9] & 83.9 [82.8, 84.7] \\
FLAME 2023 & 73.0 [70.4, 75.2] & 56.8 [53.8, 60.4] & 47.9 [45.6, 51.2] & 57.2 [54.4, 59.8] \\
FLAME 2023 (103 parameters) & 57.8 [54.4, 60.7] & 30.0 [25.7, 36.0] & 13.7 [9.4, 18.7] & 32.1 [27.3, 36.6] \\
GNM (383 parameters) & 55.0 [51.4, 59.0] & 46.3 [40.6, 50.9] & 55.0 [49.2, 58.5] & 85.3 [82.8, 87.0] \\
OVR-style rig, 72 weights, unconstrained & 88.2 [87.1, 89.1] & 89.4 [87.8, 90.5] & 92.1 [91.3, 92.7] & 97.0 [96.5, 97.3] \\
OVR-style rig, 72 weights, bounded $[0, 1]$ & 90.4 [89.5, 91.2] & 91.6 [90.4, 92.4] & 94.2 [93.7, 94.6] & 97.8 [97.5, 98.0] \\
\addlinespace[2pt]
\rowcolor{blockgrey}\multicolumn{5}{@{}l}{\textit{Five-fold neural and local models}}\\
Autoencoder$^\dagger$ & 55.1 [53.4, 56.7] & 32.4 [30.3, 35.3] & 25.0 [23.6, 26.8] & 27.0 [25.6, 28.5] \\
Weighted Autoencoder$^\dagger$ & 14.8 [13.5, 16.2] & 13.9 [13.1, 14.9] & 27.7 [26.6, 28.7] & 48.6 [46.4, 49.9] \\
Region PCA & 78.4 [76.4, 80.5] & 72.7 [69.8, 76.7] & 65.9 [63.4, 69.7] & 58.8 [57.2, 60.3] \\
SPLOCS$^\ddagger$ & 62.6 [60.0, 64.9] & 47.5 [44.8, 51.3] & 38.9 [37.1, 41.7] & 31.8 [30.8, 33.1] \\
\addlinespace[2pt]
\rowcolor{blockgrey}\multicolumn{5}{@{}l}{\textit{Mixture models}}\\
MPPCA & 12.3 [10.8, 13.7] & 2.4 [1.2, 3.5] & $-0.1$ [$-1.3$, 1.1] & 0.4 [$-0.4$, 1.2] \\
$k$-subspaces & 39.3 [37.8, 40.8] & 13.8 [12.3, 15.4] & 3.9 [2.5, 5.2] & 1.9 [1.1, 2.8] \\
\bottomrule
\end{tabularx}
\caption{\textbf{FLINCH lowers error on Subtle expressions against every listed rival.} Each cell gives how much lower FLINCH's squared error is than the rival's, in percent, with a 95\% interval in brackets. For example, 52.6 [50.3, 54.4] in the PCA row means FLINCH has 52.6\% lower squared error than PCA on Subtle expressions. Positive values favour FLINCH. Each of 20 FaMoS test people counts equally, and intervals come from resamples of people with the models unchanged. Band limits in the headers are whole-face RMS displacement from neutral, in mm. All codes use 70 parameters including the jaw unless stated. Grey rows group matched fits. PCA is principal component analysis. Log PCA gives small errors more weight during fitting. FLAME and GNM are released face models. SPLOCS learns sparse local deformations. MPPCA is a mixture of probabilistic PCA models. The OVR-style rig is a fixed headset rig. $^\dagger$Trained on 48 people plus 12 for early stopping. $^\ddagger$SPLOCS did not converge. $k$-subspaces starts from random groups of frames.}
\label{tab:s-tab1}
\end{table*}

\begin{table*}[!tp]
\centering\sffamily\fontsize{\figlabelsize}{\figtableleading}\selectfont
\setlength{\tabcolsep}{2.5pt}
\begin{tabularx}{\textwidth}{@{}Yrrrr@{}}
\toprule
Method & \shortstack{Subtle\\0.1--0.3} & \shortstack{Mild\\0.3--1} & \shortstack{Moderate\\1--3} & \shortstack{Large\\$>$3} \\
\midrule
\rowcolor{blockgrey}\multicolumn{5}{@{}l}{\textit{Ava-256 (estimated $\mu$m)}}\\
\rowcolor{flinchtint}\textbf{\name} & 74.5 & \textbf{262.5} & \textbf{515.4} & \textbf{894.6} \\
PCA & 81.5 & 283.2 & 552.6 & 918.7 \\
Log PCA & \textbf{73.8} & 263.4 & 531.0 & 949.1 \\
FLAME Open & 102.6 & 405.5 & 914.9 & 1816.7 \\
\name (FaMoS) & 91.1 & 346.2 & 769.1 & 1518.7 \\
\midrule
\rowcolor{blockgrey}\multicolumn{5}{@{}l}{FLINCH reduction against each rival (\%), 95\% intervals}\\
PCA & 16.3 [14.9, 17.8] & 14.6 [13.6, 15.5] & 14.2 [13.3, 15.1] & 5.6 [5.1, 6.2] \\
Log PCA & $-2.2$ [$-3.1$, $-1.3$] & 0.0 [$-0.9$, 1.0] & 3.7 [2.7, 4.9] & 9.8 [8.6, 11.0] \\
FLAME Open & 47.7 [45.9, 49.5] & 58.5 [57.4, 59.5] & 69.0 [68.1, 69.9] & 76.4 [75.5, 77.2] \\
\bottomrule
\end{tabularx}
\caption{\textbf{Refitted on Ava-256, FLINCH has lower error than PCA in every intensity band.} Top: root-mean-square reconstruction error in estimated micrometres, averaged equally over 176 people with frames in all four bands. Each person is scored by a fit that excludes them. Ava-256 has no calibrated metric scale, so errors are in estimated $\mu$m. Lower is better. All codes use 70 parameters including the jaw. The last row of the top block uses the libraries fitted on FaMoS, without refitting. Bottom: how much lower FLINCH's squared error is than each rival's, in percent, with 95\% intervals in brackets. Positive values favour FLINCH. Each person counts equally, and intervals come from resamples of people with the models unchanged. Orange marks the refitted FLINCH model. Bold marks the lowest error in each column. PCA is principal component analysis. Log PCA gives small errors more weight during fitting. FLAME Open is the open FLAME 2023 face model. Bands use whole-face root-mean-square displacement from neutral: Subtle 0.1--0.3, Mild 0.3--1, Moderate 1--3, Large $>3$\,mm.}
\label{tab:s-tab2}
\label{tab:s-ava-rms}
\end{table*}
\begin{table*}[!tp]
\centering\sffamily\fontsize{\figlabelsize}{\figtableleading}\selectfont
\setlength{\tabcolsep}{4pt}
\begin{tabularx}{\textwidth}{@{}Yrrrr@{}}
\toprule
Ablation & \shortstack{Subtle\\0.1--0.3} & \shortstack{Mild\\0.3--1} & \shortstack{Moderate\\1--3} & \shortstack{Large\\$>$3} \\
\midrule
2 log-spaced bands & 20.5 [19.4, 21.5] & 1.2 [0.6, 1.8] & 4.2 [3.7, 4.6] & 0.5 [0.1, 0.9] \\
3 log-spaced bands & 1.4 [1.0, 1.9] & 2.1 [1.7, 2.6] & 0.3 [0.0, 0.5] & 0.0 [$-0.1$, 0.1] \\
4 log-spaced bands & $-0.1$ [$-0.2$, 0.1] & $-0.2$ [$-0.5$, 0.1] & $-0.3$ [$-0.6$, 0.0] & 0.0 [$-0.2$, 0.2] \\
6 log-spaced bands & $-2.1$ [$-2.5$, $-1.6$] & $-0.9$ [$-1.2$, $-0.6$] & $-0.6$ [$-0.9$, $-0.3$] & 1.0 [0.6, 1.4] \\
No shared jaw & $-1.7$ [$-2.0$, $-1.5$] & $-1.7$ [$-1.9$, $-1.5$] & $-2.0$ [$-2.2$, $-1.8$] & $-2.2$ [$-2.5$, $-1.9$] \\
Library chosen by displacement & 1.0 [0.6, 1.5] & 8.6 [8.0, 9.2] & 4.4 [4.0, 4.8] & 2.5 [2.1, 2.9] \\
\bottomrule
\end{tabularx}
\caption{\textbf{Four bands lower error on Subtle expressions compared with two bands.} Each cell gives how much lower the squared error of FLINCH with four bands is than each ablation's, in percent, at 70 parameters, with a 95\% interval in brackets. Positive values favour four bands. Each of the 75 FaMoS people outside the test set is scored by a fit that excludes them, and each person counts equally. Intervals come from resamples of people with the models unchanged. Log-spaced bands replace the rounded band limits with exactly log-spaced limits. Library chosen by displacement picks the library from the frame's measured displacement instead of the lowest fit error. No shared jaw removes the shared jaw directions. Bands use whole-face root-mean-square displacement from neutral: Subtle 0.1--0.3, Mild 0.3--1, Moderate 1--3, Large $>3$\,mm.}
\label{tab:s-ablate}
\end{table*}
\paragraph{Published spaces and mixtures.}
\label{sec:s-mix}
At 70 parameters, \name has 12.3\% lower squared error on Subtle expressions than MPPCA, 8.2\% lower than MPPCA fitted without means and 39.3\% lower than $k$-subspaces started from random groups. Given intensity groups to start from, $k$-subspaces comes within 2.9\%, interval $[1.4, 4.4]$. Intensity is therefore an effective grouping, and \name builds it in directly, without fitting a mixture (main-paper Sec.~\ref{sec:main}). Both MPPCA versions are fitted with the same weighting on all 75 FaMoS people outside the test set. FLAME and GNM are used as linear expression models with the same shared jaw, without their pose-dependent corrections. GNM uses its released 383 parameters. PCA fitted on FaMoS has 41 to 57\% lower squared error than FLAME 2023, which is learned from other 3D scan data. Comparing with PCA fitted on the same data therefore shows the effect of the representation alone.

\paragraph{Autoencoders and headset rig.}
\label{sec:s-ae}
On Subtle expressions, \name has 55.1\% lower squared error than the ordinary autoencoder and 14.8\% lower than the autoencoder weighted by intensity. \Cref{tab:s-tab1} gives the 95\% intervals. Both autoencoders use one architecture with 173,190 weights and a 70-parameter code.
\label{sec:s-rig}
\name lowers error against the OVR-style VR headset rig in all four bands (\cref{tab:s-tab1}). The rig has 72 parameters, each a named blendshape weight. With each weight kept between 0 and 1, as in a headset, its RMS errors from Subtle to Large are 192.4, 523.0, 1113.0 and 3213.2\,$\mu$m. \Cref{tab:s-main-rms} reports the rig without this limit. These fits measure how much of each expression the rig can represent. Recorded headset signals are evaluated in \cref{sec:s-headset-events}.

\paragraph{Ava-256 refitting.}
\label{sec:s-ext}
Refitted on Ava-256, \name has 16.3\% lower squared error than PCA on Subtle expressions, with a 95\% interval $[14.9, 17.8]$. Log PCA has 2.2\% lower squared error than \name on Subtle expressions here (main-paper Sec.~\ref{sec:transfer}), and \name has lower error than Log PCA on Moderate and Large expressions. \Cref{tab:s-tab2} gives the paired comparisons with PCA, Log PCA and FLAME Open. Like any learned basis, libraries fitted on FaMoS are best refitted for a new capture setup. Even without refitting, they have lower error than FLAME Open in every band (\cref{tab:s-tab2}). Without refitting, \name fitted on FaMoS also lowers squared error on Subtle expressions by 1.3\% on D3DFACS and 8.5\% on CoMA, two further 3D expression datasets.

\paragraph{Bands, selection and jaw.}
\label{sec:s-variants}
\label{sec:s-ablate}
Four bands give 20.5\% lower squared error than two bands on Subtle expressions, with a 95\% interval $[19.4, 21.5]$. Four bands give almost all of the gain. Six bands lower Subtle error by a further 2.1\% but raise error on Large expressions by 1.0\% (main-paper Sec.~\ref{sec:transfer}). Choosing the library by fit, rather than by measured band, lowers error on Mild expressions by 8.6\%. The three shared jaw directions keep jaw control identical in every library (\cref{tab:s-ablate}).

\paragraph{Switching libraries.}
\label{sec:s-hyst}
Over time, \name switches library only when another one fits better by a fixed margin of 0.066 estimated mm$^2$ per vertex, the 75th percentile of the positive switching gains on training frames. This cuts library switches on Mild Ava-256 test frames from 8.1\% to 0.02\%. Frame-to-frame error, the RMS change of the reconstruction error between adjacent frames, falls from 5\% above PCA's to 6\% below (main-paper Sec.~\ref{sec:transfer}).

\paragraph{FLINCH-4D benchmark.}
\label{sec:s-benchmark}
\Cref{tab:s-natural} gives benchmark errors on the 42-person test split. FLINCH-4D labels and the scripts to rebuild the benchmark from the original Ava-256 data will be released (main-paper Sec.~\ref{sec:data}).
\begin{table}[!htbp]
\centering\sffamily\fontsize{\figlabelsize}{\figtableleading}\selectfont\setlength{\tabcolsep}{3pt}
\begin{tabularx}{\columnwidth}{@{}Yrrrr@{}}\toprule
Method & \shortstack{Subtle\\0.1--0.3} & \shortstack{Mild\\0.3--1} & \shortstack{Moderate\\1--3} & \shortstack{Large\\$>$3} \\\midrule
\rowcolor{flinchtint}\textbf{FLINCH-Ava} & 0.071 & \textbf{0.270} & \textbf{0.531} & \textbf{1.110} \\
PCA & 0.076 & 0.290 & 0.568 & 1.126 \\
Log PCA & \textbf{0.070} & 0.275 & 0.554 & 1.229 \\
FLAME Open & -- & 0.401 & 0.926 & 2.333 \\
Autoencoder & 0.083 & 0.287 & 0.566 & 1.147 \\
\bottomrule\end{tabularx}
\caption{\textbf{FLINCH-Ava has the lowest benchmark error on Mild to Large expressions.} Entries are root-mean-square reconstruction errors in estimated millimetres, averaged equally over the people in the 42-person FLINCH-4D test split. Lower is better. All codes use 70 parameters including the jaw. FLINCH-Ava is FLINCH fitted on the FLINCH-4D training split. Orange marks it and bold marks each minimum. PCA is principal component analysis. Log PCA gives small errors more weight during fitting. FLAME Open is the open FLAME 2023 face model. A dash marks a value not reported in main-paper Sec.~\ref{sec:data}. Bands use whole-face root-mean-square displacement from neutral: Subtle 0.1--0.3, Mild 0.3--1, Moderate 1--3, Large $>3$\,mm.}
\label{tab:s-natural}
\end{table}\par
\section{Expression preservation and coding}
\label{sec:s-meaning}
\paragraph{Facial actions.}
\label{sec:s-exprpres}
On FaMoS, \name keeps 99\% of the Subtle eyelid peak, with a 95\% interval $[98, 100]$, against 94\% $[92, 96]$ for PCA and 86\% $[79, 91]$ for FLAME 2023. Peak retention is the reconstructed peak divided by the original peak, so values above 100\% overshoot. On Ava-256 regional Subtle eye peaks, the retained shares are 85.6\% $[83.8, 87.2]$ for \name, 79.5\% $[77.0, 81.9]$ for PCA, 86.2\% $[84.4, 87.9]$ for Log PCA and 65.2\% $[62.3, 68.0]$ for FLAME 2023. These are 98 Subtle eye events from 89 people, placed in bands by the eye region's own displacement, with each person scored by a fit that excludes them.

\begin{table}[!htbp]
\centering\sffamily\fontsize{\figlabelsize}{\figtableleading}\selectfont
\setlength{\tabcolsep}{2.5pt}
\begin{tabularx}{\columnwidth}{@{}Yrrrr@{}}
\toprule
Method & \shortstack{Subtle\\0.1--0.3} & \shortstack{Mild\\0.3--1} & \shortstack{Moderate\\1--3} & \shortstack{Large\\$>$3} \\
\midrule
\rowcolor{flinchtint}\textbf{\name} & 0.44 & \textbf{3.22} & \textbf{6.66} & \textbf{0.04} \\
PCA & 0.79 & 3.47 & 8.71 & 1.69 \\
Log PCA & 0.35 & 4.36 & 12.08 & 5.45 \\
FLAME 2023 & 0.88 & 4.29 & 17.13 & 15.02 \\
FLAME 2020 & \textbf{\textminus0.18} & 4.61 & 19.11 & 24.15 \\
\bottomrule
\end{tabularx}
\caption{\textbf{FLINCH loses the least classifier accuracy on Mild, Moderate and Large expressions.} Each cell is the accuracy lost when an expression classifier trained on original meshes is tested on reconstructions, in percentage points, at 70 parameters on 20 FaMoS test people. Lower is better, and a negative value means accuracy rises. On original meshes the classifier reaches 3.35, 18.89, 39.43 and 58.00\% from Subtle to Large. Chance level is 3.85\% (26 expression classes), so the classifier cannot separate methods on Subtle expressions, where peak retention is used instead. Orange marks FLINCH. Bold marks each minimum. PCA is principal component analysis. Log PCA gives small errors more weight during fitting. FLAME 2023 and 2020 are released face models. Bands use whole-face root-mean-square displacement from neutral: Subtle 0.1--0.3, Mild 0.3--1, Moderate 1--3, Large $>3$\,mm.}
\label{tab:s-classifier}
\end{table}

\paragraph{Classifier by band.}
\label{sec:s-classifier}
A classifier trained on original meshes reaches 43.6\% accuracy across all frames. It loses only 3.1 points on \name reconstructions, against 4.7 on PCA and 13.8 on FLAME 2023 (main-paper Fig.~\ref{fig:fidelity}). By band, \name loses the least accuracy on Mild, Moderate and Large expressions: 3.22, 6.66 and 0.04 points (\cref{tab:s-classifier}). On Subtle expressions the classifier is near chance even on the original meshes, so that band cannot separate the methods. Subtle preservation is measured directly by peak retention, where \name keeps 99\% of the eyelid peak.

\paragraph{Library sizes.}
\label{sec:s-adaptive}
Giving each library its own number of parameters saves a further 9 to 14\% of bits on Subtle to Moderate expressions (main-paper Sec.~\ref{sec:exprrec}). On the FLINCH-4D test split, the selected libraries use 51, 51, 59 and 70 parameters from Subtle to Large. Each frame uses the number of parameters of the library that fits it best. At 8 bits per parameter plus the 2-bit index, the code is 14.2\%, 10.8\%, 9.3\% and 2.2\% shorter than the 560 bits of a single basis from Subtle to Large, with 95\% intervals $[11.6, 16.9]$, $[9.1, 12.4]$, $[8.3, 10.3]$ and $[1.7, 2.8]$. Counting 32 bits per parameter gives 14.5\%, 11.1\%, 9.6\% and 2.5\% $[2.0, 3.1]$. On Large expressions, error stays on par, with a ratio of mean squared errors of 0.994 $[0.985, 1.002]$. Library sizes were chosen on validation people from 20,736 combinations. Errors are measured before rounding to bits.

\paragraph{Mixed expressions.}
\label{sec:s-mixed}
When the lower-face expression is much larger than the upper-face expression, the encoder may favour the library that fits the upper face (main-paper Sec.~\ref{sec:exprrec}). At the same 562 bits, this cuts upper-face squared error, measured relative to the original upper-face movement, from 0.167 $[0.143, 0.193]$ to 0.116 $[0.100, 0.134]$ on 17 people, a 30\% reduction. Whole-face error over 20 people rises by only 0.02\%, interval $[0.01, 0.03]$. The encoder keeps this rise below 1\% over each full recording.

\paragraph{Bits and decoding cost.}
\label{sec:s-cost}
At 70 parameters, \name stores 271 basis columns: four libraries of 67 directions and 3 shared jaw directions. These columns are stored once, 5.8\,MB as 32-bit numbers on the 1,787 face vertices. Decoding remains one matrix multiplication, as with a single basis (main-paper Sec.~\ref{sec:encoding}). Per frame, the 2-bit index selects one library, and its 67 columns with the 3 jaw columns form a $5{,}361\times70$ matrix (1,787 vertices, 3 coordinates each). Multiplying this matrix by the 70 parameters takes 375,270 multiplications, exactly as many as PCA at 70 parameters. At matched error on Subtle expressions, PCA needs 910 bits per frame (91 parameters at 10 bits, with rounding error included, whereas the 117-parameter match in \cref{sec:s-results} is measured before rounding), against 562 for \name, a saving of 38.2\%, 95\% interval $[36.1, 39.6]$ (main-paper Sec.~\ref{sec:exprrec}). Each library uses its own rounding ranges, and rates come from a fixed grid of bit counts. The saving combines the libraries and their rounding ranges. On Large expressions, the error ratio at equal bits is 0.99.

\begin{figure*}[!t]
\centering

\includegraphics[width=\textwidth]{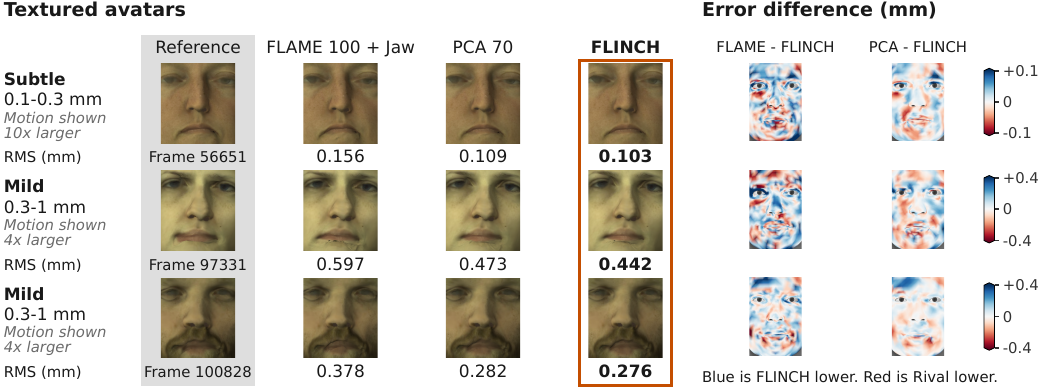}
\caption{\textbf{FLINCH has lower error at the peak frame in these 3 textured avatars.} The differences are below a millimetre, so the textured faces look alike. The numbers and the difference maps show where FLINCH is more accurate. Rows show 3 Ava-256 people excluded from fitting. Grey panels show original motion and orange marks FLINCH. Displays magnify motion 10, 4 and 4 times so that it is visible, which also exaggerates the face shape. All errors use unmagnified motion. Numbers are the peak frame's root-mean-square (RMS) error in estimated mm. Lower is better. Difference maps show the rival's error minus FLINCH's in estimated mm: blue favours FLINCH, red favours the rival, and white means equal error. Rows were chosen by a fixed rule before any comparison was seen: the people at the 75th (Subtle), 25th (Mild) and 50th (Mild) percentiles of the ratio of FLAME to FLINCH error among people with textures. PCA (principal component analysis) and FLINCH use 70 parameters including jaw. FLAME Open is the open FLAME 2023 face model with 100 expression directions plus jaw. Bands use whole-face root-mean-square displacement from neutral: Subtle 0.1--0.3, Mild 0.3--1, Moderate 1--3, Large $>3$\,mm. Bold marks the lowest error in each row. Colour scales differ per row. FLAME 100 + jaw is FLAME Open with all 100 expression components plus the jaw (103 parameters).}
\label{fig:s-avatar}
\end{figure*}
\paragraph{Textured avatars.}
\label{sec:s-avatar}
\Cref{fig:s-avatar} shows 3 Ava-256 people excluded from fitting, with the same static texture for every method.

\section{Real headset signals, image input and 4D scans}
\label{sec:s-inputs}
\subsection{Recorded headset signals}
\label{sec:s-headset-events}
BlendME~\cite{blendme2026} records micro-expressions in headset recordings, each annotated as a short event. We pass the 76 recorded micro-expression events of 20 people through the OVR-style VR headset rig and code the result at 70 parameters. Peak retention is the coded peak of an event divided by its recorded peak. \name keeps 87.9\% of each event's peak on average. This is 1.15 percentage points more than PCA, 95\% interval $[0.31, 2.46]$, and on par with FLAME 2023, with a difference of 0.44 $[-0.86, 1.98]$ points. These are the results reported in main-paper Sec.~\ref{sec:exprrec}.

\subsection{Rendered-image prediction}
\label{sec:s-tracker}
A tracker trained on rendered Ava-256 meshes predicts expression parameters from images. It has one output head each for PCA, \name and FLAME parameters, trained on the same images. On 20 people held out from tracker training, it drives \name within 0.3\% of PCA's error in every band and below FLAME's error in all four, as reported in main-paper Sec.~\ref{sec:exprrec}. Tracker training details are in \cref{tab:s-baselines}.
\label{fig:s-tracker}\label{fig:s-tolerance}\label{tab:s-tracker}\label{sec:s-tolerance}

\subsection{Spontaneous micro-expressions in 4D}
\label{sec:s-4dme}
4DME~\cite{fourDME2023} records spontaneous micro-expressions as 3D face scans over time. On all 22 analysable clips from 12 people in 4DME, \name has 2\% lower root-mean-square (RMS) error than PCA and 9\% lower than FLAME 2023 over each annotated facial-action region, with 95\% intervals above zero for both (\cref{tab:s-4dme}). All three use 70 parameters, and the libraries are not refitted on 4DME. Their errors are 0.153, 0.155 and 0.167\,mm. \name has lower error than FLAME 2023 for all 12 people.
\begin{table}[!htbp]
\centering\sffamily\fontsize{\figlabelsize}{\figtableleading}\selectfont\setlength{\tabcolsep}{3pt}
\begin{tabularx}{\columnwidth}{@{}Yr@{}}
\toprule
Method & \shortstack[r]{Lower RMS error with FLINCH ($\mu$m)\\95\% interval} \\
\midrule
PCA (70) & \textbf{2.8} [0.9, 5.0] \\
FLAME 2023 (70) & \textbf{14.7} [9.3, 21.3] \\
\bottomrule
\end{tabularx}
\caption{\textbf{FLINCH has lower error than PCA and FLAME 2023 on spontaneous 4DME micro-expressions.} Each entry is how much lower FLINCH's root-mean-square error is than the rival's, in micrometres ($\mu$m), over each annotated facial-action region. Positive values favour FLINCH. All models use 70 parameters including the jaw, and the libraries are not refitted on 4DME. Events are averaged within each person, and people then count equally. Brackets give 95\% intervals from resampling people with the models unchanged. Both rows use all 22 analysable clips from 12 people. PCA means principal component analysis. FLAME 2023 is the released face model. 4DME records 3D face scans of micro-expressions over time. Bold marks reductions whose 95\% interval is above zero.}
\label{tab:s-4dme}
\end{table}

\section{Ethics and release}
\label{sec:s-ethics}
\paragraph{Ethics and consent.}
Ethics approval was obtained. Ava-256 records written consent for capture and distribution and is used under CC BY-NC 4.0. No identifiable faces from publication-restricted data are shown.
FaMoS is used under its non-commercial research licence. The BlendME paper is published~\cite{blendme2026}, and its headset recordings are available on request from the dataset authors for research use, with participant consent.

\paragraph{Release.}
FLINCH-4D labels, splits, masks, evaluation code, FLINCH-Ava libraries, per-frame displacements and scripts that rebuild them from Ava-256 will be released under Ava-256's CC BY-NC 4.0 terms, with FLAME Open attribution. Raw Ava-256 meshes, images and textures are not redistributed, and no data derived from FaMoS is included. Audio and person-specific neutral geometry are also excluded. Third-party weights and portraits are not bundled.

\end{document}